%% file: main.tex
\documentclass[12pt, a4paper, oneside]{Thesis} 
\usepackage{wrapfig}
\usepackage{lscape}
\usepackage{rotating}
\usepackage{graphicx}
\usepackage{caption}
\usepackage{amsmath}
\usepackage{multirow}
\usepackage{listings}

\usepackage{lineno,hyperref}
\modulolinenumbers[5]

\usepackage{amssymb}
\usepackage{graphicx}
\usepackage{array}
\usepackage{float}
\usepackage{placeins}
\usepackage{stackengine}
\usepackage{url}
\usepackage{numprint}
\usepackage{caption}

\usepackage{booktabs}  
\usepackage{siunitx}
\usepackage{subfigure}

\nprounddigits{3}
\newcolumntype{P}[1]{>{\centering\arraybackslash}p{#1}}
\newcolumntype{M}[1]{>{\centering\arraybackslash}m{#1}}

\setstackEOL{\#}
\setstackgap{L}{12pt}

\graphicspath{{Pictures/}} 
\usepackage[square,sort,comma,numbers]{natbib} 

\hypersetup{urlcolor=black, colorlinks=false} 

\thesistitle{Model Blending for Text Classification}
\supervisor{Prof. Jayanta Mukhopadhya, Dr Sunav Choudhary, Dr Vishwa Vinay}
\degree{Master of Science}
\degreemajor{Mathematics and Computing}
\authors{Ramit Pahwa}
\rollno{14MA20029}
\university{Indian Institute of Technology Kharagpur}
\department{Department of Computer Science}
\unisite{http://www.iitkgp.ac.in}
\depsite{http://www.iitkgp.ac.in/department/CE}
\placeshrt{Kharagpur}
\placelng{Kharagpur - 721302, India}
\datesub{April 29, 2018}
\datesig{April 29, 2018}
\semsub{Autumn Semester, 2018-19}
\coursecd{Project-II (MA57012) }

\title{\ttitle} 
\begin{document}


\frontmatter 

\setstretch{1.6} 

\fancyhead{} 
\rhead{\thepage} 
\lhead{} 

\newcommand{\HRule}{\rule{\linewidth}{0.5mm}} 

\hypersetup{pdftitle={\ttitle}}
\hypersetup{pdfsubject=\subjectname}
\hypersetup{pdfauthor=\authornames}
\hypersetup{pdfkeywords=\keywordnames}

\maketitle

\clearpage 


\Declaration


\addtotoc{Certificate} 

\certificate{\addtocontents{toc}{} 
\clearpage 

\acknowledgements 

I express my deep sense of gratitude to my guide Prof. Jayanta Mukhopadhya
for her valuable guidance and inspiration throughout the course of this work. I am
thankful to her for her help, guidance, directions and support and his motivation
towards independent thinking. It has been a great experience working under her in
the cordial environment.I would i also like to thank Prof. Dejani Chakraborty, Department of Computer Science,IIT Kharagpur,Dr. Viswa Vinay and Dr. Sunav Choudhary, Adobe Research for their constant guidance and support.
\clearpage 


\addtotoc{Abstract} 

\abstract{\addtocontents{toc}{} 

Deep neural networks (DNNs) have proven successful in a wide variety of applications such as speech recognition and synthesis, computer vision, machine translation, and game playing, to name but a few. However, existing deep neural network models
are computationally expensive and memory intensive, hindering
their deployment in devices with low memory resources or in
applications with strict latency requirements. Therefore, a natural
thought is to perform model compression and acceleration
in deep networks without significantly decreasing the model
performance, which is what we call \textbf{reducing the complexity}. In the following work, we try reducing the complexity of state of the art LSTM models for natural language tasks such as text classification, by distilling their knowledge to CNN based models, thus reducing the inference time(or latency) during testing.
}

\clearpage 

\pagestyle{fancy} 

\lhead{\emph{Contents}} 
\tableofcontents 

\lhead{\emph{List of Figures}} 
\listoffigures 

\lhead{\emph{List of Tables}} 
\listoftables 
\lstlistoflistings
%
%
%
%


\mainmatter 

\pagestyle{fancy} 


\input{Chapters/Chapter1}

\input{Chapters/Chapter2}

\input{Chapters/Chapter3} 
\input{Chapters/Chapter4}
\input{Chapters/Chapter5}

\input{Chapters/Chapter6}
\input{Appendices/AppendixA}

\label{Bibliography}

\lhead{\emph{Bibliography}} 

\bibliographystyle{splncs04}
\bibliography{Bibliography} 

\end{document}

%% file: Chapters/Chapter1.tex

\chapter{Introduction} 

\label{Chapter 1} 


\lhead{Chapter 1. \emph{Introduction}} 

\section{Motivation}

\paragraph{Reducing Complexity of Deep Neural Networks - Why ?}
Deep neural networks (DNNs) have proven successful in a wide variety of applications such as speech recognition and synthesis, computer vision, machine translation, and game playing, to name but a few. However, existing deep neural network models
are computationally expensive and memory intensive, hindering
their deployment in devices with low memory resources or in
applications with strict latency requirements. Therefore, a natural
thought is to perform model compression and acceleration
in deep networks without significantly decreasing the model
performance, which is what we call \textbf{reducing the complexity}. As larger neural networks with more layers and nodes
are considered, reducing their complexity
becomes critical, especially for some real-time applications
such as online learning and incremental learning. In addition,
recent years witnessed significant progress in virtual
reality, augmented reality, and smart wearable devices, creating
unprecedented opportunities for researchers to tackle
fundamental challenges in deploying deep learning systems to
portable devices with limited resources (e.g. memory, CPU,
energy, bandwidth).

\section{Arriving at the Problem}

\paragraph{Related Work}

At the high level, there are two research directions for reducing model complexity: one is to introduce a completely new architecture with reduced complexity which mimics the original model and other is to make changes to the original model itself, without altering its overall structure. By not altering the structure, we mean applying techniques like quantization, pruning, huffman coding to the model we want to compress. Here again there are 2 directions: one is to compress while training, for example training a model with weights that are quantized, like in trained ternary quantization by \cite{zhu2016trained}, and other is compressing a pre-trained model, for example using quantization, pruning etc. like in deep compression by \cite{han2015deep}.\\

The second approach is to introduce a new architecture that mimics the original model. Here one idea is knowledge distillation due to \cite{hinton2015distilling}, which is basically to make the student model learn the behaviour of the teacher model. The caveat here is that the the architecture of the student model has to be hand designed. Due to this, there have been explorations on learning the `optimal' architecture so to speak. There have been some very recent works on architecture search, which try to search the architecture from scratch via reinforcement learning or genetic algorithms, with the downside being it is computationally expensive. Another approach could be restricting the search space to the teacher model only. It relies on the premise that since the teacher model is able to achieve high accuracy on the dataset, it already contains the components required to solve the task and therefore is a suitable search space for compressed architecture - an idea that has been used in N2N compression by \cite{ashok2017n2n}. There also have been some very recent works that try to combine some of these approaches. For example, quantized distillation by \cite{polino2018model} which is basically knowledge distillation but with the weights quantized while training to learn the behaviour of the teacher model.

\paragraph{Research gaps}
Much of the recent work on reducing model complexity is focused on learning new architectures from scratch or using a teacher model, via reinforcement learning, which is the second approach mentioned above. However, the reward functions that these reinforcement learning methods use are based on just accuracy and  compression. None of these methods incorporate latency in their reward function, which can be important in latency strict applications. We first thought of working in this direction, but the reinforcement learning paradigm for architecture search requires a huge amount of computational resources, which we didn't have at our disposal. Thus we thought of reducing the scope of the problem, and during literature reading came across a paper by \cite{bai2018empirical} that compared RNNs and CNNs for sequence modelling. It said that CNNs are advantageous over RNNs for sequence modelling in that parallel processing is possible for them, resulting in reduced latency during training and testing. Since almost all state of the art models for sequence modelling tasks like machine translation etc. are RNN based, we thought of working on distilling the knowledge from these architectures to CNN based models, a recent study proposes utilizing the outputs of LSTM while training the CNN which tries to incorporate 'dark knowledge' \cite{hinton2015distilling} for automatic speech
recognition task \cite{geras2015blending}.
We propose blending for text classification task.

\section{Text Classification}
Text classification is fundamental problem in Natural Language Processing(NLP) and is an essential component in applications such as sentiment analysis \cite{maas-EtAl:2011:ACL-HLT2011}, question classification \cite{voorhees1999trec} and topic classification \cite{zhang2015character}.In recent year there has been a shift from traditional bag of words (BoW) model, which treats textual data as an unordered set of words \cite{wang2012baselines} to more recent order sensitive models which have achieved exceptional performance across numerous NLP tasks.
The challenge for textual modelling is to tackle how to capture features from various textual units such as phrases, sentences and documents.
Benefitting from the sequential nature of the textual data, the recurrent neural network (RNN) in particular Long short term memory (LSTM) \cite{zhou2016text} units has been successful in modelling numerous classification tasks.
But it comes with its limitations as it processes the data in a sequential manner, thereby increasing the execution latency and hindering its deployment in strict latency environments.
The alternative deep learning architecture which in the recent years has gained popularity over LSTM is the convolution neural network (CNN) \cite{kim2014convolutional} primarily due to its parallelism, flexible receptive filter window \cite{Wang2018Densely} and low memory requirement while training in comparison to LSTM network. The inherent difference in the architectures of CNN and LSTM pose an interesting question of whether the functions they represent are different and if so, is there a method to distill knowledge from one of them to other.
In this paper we apply \textbf{model blending}, a method to train very accurate CNN models by using LSTM priors to guide the training process, to text classification. The choice to distill the knowledge from LSTM to CNN and not vice-versa is motivated by the fact that CNN's are faster during inference time due to their parallelism.
We thus arrive at the problem statement.

\section{Problem Statement}

Distilling knowledge from state of the art RNN architectures \cite{} for natural language tasks such as text classification, machine translation etc. into CNN based architectures and comparing the latency and accuracy. As well as effectively employ a Computer Vision (CV) like transfer learning mechanism to Natural Language Processing (NLP) . 
In summary, the contributions to achieve this are :
\begin{itemize}
    \item We propose blending into CNN using state of the art LSTM priors resulting in accurate CNN classifiers which are 15x more computationally efficient at test time and take a fraction of the training time as compared with the training of LSTM network.

    \item We empirically find that using a more accurate Teacher LSTM we can improve on the performance of the model. 
    \item We extensively evaluate the blending process across text classification tasks such sentiment analysis \cite{maas-EtAl:2011:ACL-HLT2011}, question classification \cite{voorhees1999trec} and topic classification and obtain competitive performance when compared with state of the art models.
    \item We provide vision style transfer learning by using recurrent language model trained on the corpus\cite{howard2018universal}, instead of traditional embedding based transfer \cite{pennington2014glove}.
\end{itemize}

%% file: Chapters/Chapter2.tex

\chapter{RNNs vs CNNs} 

\label{Chapter 2} 


\lhead{Chapter 2. \emph{RNNs vs CNNs: Which is better for Sequence Modelling ?}} 

\section{RNNs vs CNNs}

Deep learning practitioners commonly regard recurrent architectures
as the default starting point for sequence modeling
tasks.  On the other hand, recent research indicates that certain convolutional
architectures can reach state-of-the-art accuracy
in audio synthesis, word-level language modeling, and machine
translation (\cite{oord2016wavenet}; \cite{kalchbrenner2016neural}; \cite{dauphin2016language}; \cite{gehring2016convolutional}).
This raises the question of whether these successes of convolutional
sequence modeling are confined to specific application
domains or whether a broader reconsideration of
the association between sequence processing and recurrent
networks is in order. The paper by \cite{bai2018empirical} addresses this question by conducting a systematic empirical
evaluation of convolutional and recurrent architectures
on a broad range of sequence modeling tasks. We provide a summary of the advantages and disadvantages of using CNNs over RNNs for sequence modelling, as presented in the paper.

\paragraph{Advantages of using CNNs over RNNs}

\begin{itemize}
    \item \textbf{Parallelism}: Unlike in RNNs where the predictions for
later timesteps must wait for their predecessors to complete,
convolutions can be done in parallel. Therefore, in both training and
evaluation, a long input sequence can be processed as a
whole in CNN, instead of sequentially as in RNN.

    \item \textbf{Flexible receptive field size}: A CNN can change its receptive
field size in multiple ways. For instance, stacking
more dilated (causal) convolutional layers, using larger
dilation factors, or increasing the filter size are all viable
options (with possibly different interpretations). CNNs
thus afford better control of the model’s memory size,
and are easy to adapt to different domains.

\item  \textbf{Stable gradients}: Unlike recurrent architectures, CNN
has a backpropagation path different from the temporal
direction of the sequence. CNN thus avoids the problem for RNNs.

\item \textbf{Low memory requirement for training}: Especially in
the case of a long input sequence, RNNs and GRUs can
easily use up a lot of memory to store the partial results
for their multiple cell gates. However, in a CNNs the filters
are shared across a layer, with the backpropagation path
depending only on network depth. Therefore in practice, gated RNNs likely to use up to a multiplicative
factor more memory than CNNs.

\item \textbf{Variable length inputs}: Just like RNNs, which model
inputs with variable lengths in a recurrent way, CNNs
can also take in inputs of arbitrary lengths by sliding the
1D convolutional kernels. This means that CNNs can be
adopted as drop-in replacements for RNNs for sequential
data of arbitrary length.

\end{itemize}

\paragraph{Disadvantages of using CNNs over RNNs}

\begin{itemize}
    \item  \textbf{Data storage during evaluation}: In evaluation/testing,
RNNs only need to maintain a hidden state and take in a
current input \(x_t\) in order to generate a prediction. In other
words, a ``summary” of the entire history is provided by
the fixed-length set of vectors \(h_t\), and the actual observed
sequence can be discarded. In contrast, CNNs need to
take in the raw sequence up to the effective history length,
thus possibly requiring more memory during evaluation.

    \item \textbf{Potential parameter change for a transfer of domain}: 
Different domains can have different requirements on the
amount of history the model needs in order to predict.
Therefore, when transferring a model from a domain
where only little memory is needed
to a domain where much longer memory is required, CNN may perform poorly for not
having a sufficiently large receptive field.

\end{itemize}

\paragraph{Conclusion}

The comparision above demonstrates that CNNs may well be the best for sequence modelling in certain situations. The advantage that we are most concerned about is the reduced inference time due to parallelism. In the next chapter, we introduce a method for distilling the knowledge of RNN architectures into CNN architectures.

%% file: Chapters/Chapter3.tex

\chapter{Methodology} 

\label{Chapter 3} 


\lhead{Chapter 3. \emph{Blending RNNs into CNNs}} 
Although LSTMs yield state of the art performance on numerous tasks, they are slow to evaluate at test time, thus restricting their deployment in strict latency environments.
CNNs on the other hand are much faster at test time on account of their parallelism, though subpar in performance as compared to LSTMs. The proposed system aims to improve the performance of the CNNs while reaping the benefits we get from it during deployment.
The methodology, which we call \emph{model blending}, is inspired by \cite{geras2015blending}. Their objective is to improve the performance of the CNN model on automatic speech recognition task by mimicking the output of the LSTM model.
The process is somewhat analogous to knowledge distillation \cite{hinton2015distilling}, in which a bigger more powerful model referred as the teacher is used to train a compact model referred as the student.
The teacher here is a variant of Recurrent Neural Network i.e Long Short-Term Memory (LSTM) which can capture capture long term dependencies and the student is a Convolutional Neural Network (CNN). The choice of having the CNN model as the student is what gives us benefits in terms of latency at test time.  Thus, the blending process gives the benefits of ensembling the predictions of both the models (namely LSTM and CNN) while bypassing the usual objection of an ensemble requiring too much computation at test time. An intuitive implementation is shown in figure 1.
\begin{figure}[h]

\includegraphics[scale=0.7]{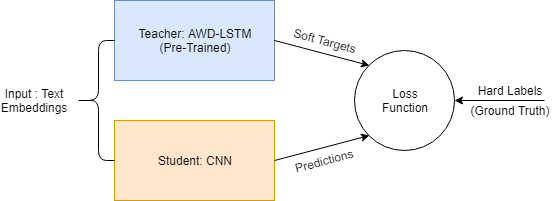}
\caption{Blending Process} 
\end{figure}
\section{Teacher: LSTM}
LSTMs exhibit superior performance not only in classification \cite{merity2017regularizing,howard2018universal} and speech , but also in handwriting recognition \cite{graves2009offline,graves2013generating} and parsing \cite{vinyals2015grammar}, this is primarily due their ability to capture long range interaction which is an essential component in understanding the semantics of the sentence. The Figure shows the LSTM unit.
The governing equations are as follows:
\begin{equation}
f_t = \sigma_g(W_{f} x_t + U_{f} h_{t-1} + b_f) 
\end{equation}
\begin{equation}
i_t = \sigma_g(W_{i} x_t + U_{i} h_{t-1} + b_i) 
\end{equation}
\begin{equation}
o_t = \sigma_g(W_{o} x_t + U_{o} h_{t-1} + b_o) 
\end{equation}
\begin{equation}
c_t = f_t \circ c_{t-1} + i_t \circ \sigma_c(W_{c} x_t + U_{c} h_{t-1} + b_c) 
\end{equation}
\begin{equation}
h_t = o_t \circ \sigma_h(c_t)
\end{equation}
\begin{figure}[h]

\includegraphics[scale=0.6]{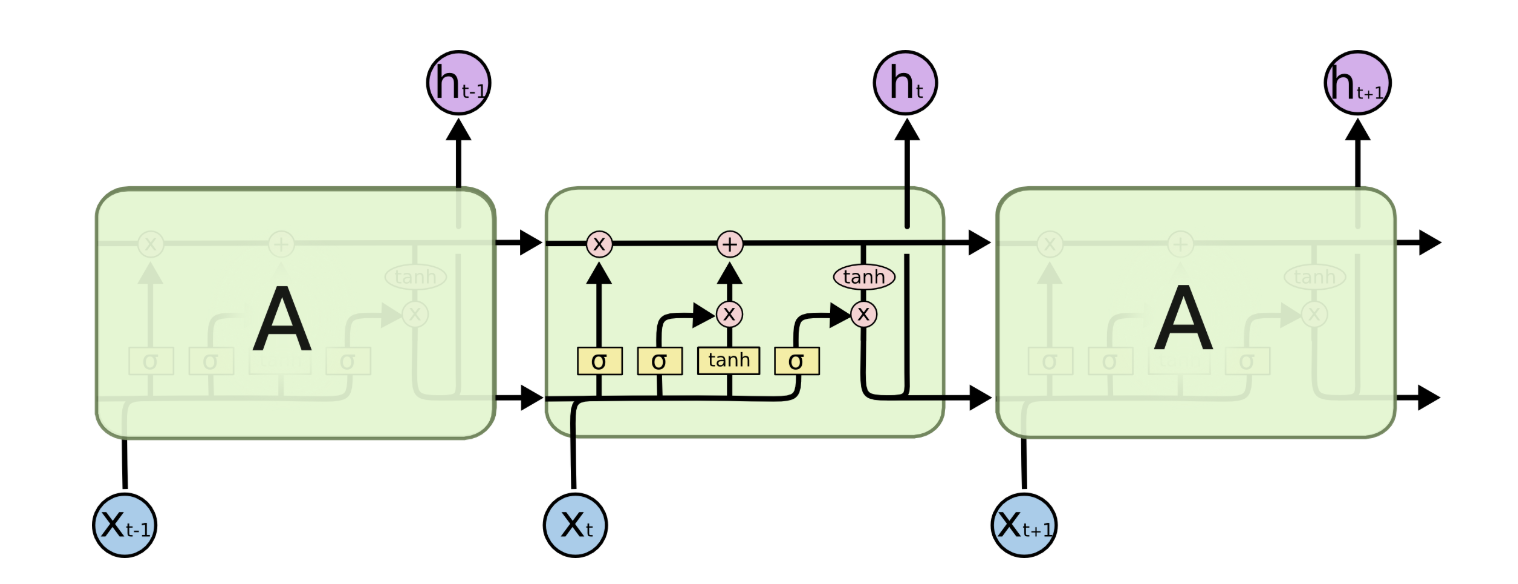}
\caption{The repeating module in an LSTM contains four interacting layers.} 
\end{figure}
For classification tasks, we train two networks, the first one is a Bi-Directional LSTM network \cite{zhou2016text} which is a tough baseline to break for most NLP tasks.
The sentence $S = (x_{1}, x_{2}, ..., x_{m})$, where $m$ is the length of the sentence is taken as input and then we use glove embedding \cite{pennington2014glove} of size 300 to embed the words of the sentence.
The above obtained sentence matrix $S$ has a dimension of $N \times{d}$, here $N$ is the batch size and $d$ is the fixed embedding dimension of each word.
This matrix is given as input to the Bi-direction LSTM network, the last hidden state $ \Vec{h_{t}} $ of the forward network along with the last hidden state of the backward network $ \overleftarrow{h_{t}}$ is passed into the fully connected network to predict the outputs.
This forms a baseline RNN teacher for our experiments. We further utilize the state of the art architecture i.e AWD-LSTM \cite{merity2017regularizing} architecture and employ the same training regime mentioned in \cite{howard2018universal} namely discriminative fine-tuning, gradual unfreezing of layers and slanted triangular learning rates to achieve state of the art performance for these tasks.The Pseudo code is provided below to define the RNN network in pytorch.
\begin{lstlisting}[language=Python, caption=Algorithm for Recurrent Neural Network LSTM applied to Text Data, numberstyle=\tiny]
import torch.nn as nn

class RNN(nn.Module):
    def __init__(self, v_s, emb_dim, h_dim, o_dim, n, bi, d):
        super().__init__()
        
        self.embedding = nn.Embedding(v_s, emb_dim)
        self.rnn = nn.LSTM(emb_dim, h_dim, n, bi, dropout=d)
        self.fc = nn.Linear(h_dim*2, o_dim)
        self.dropout = nn.Dropout(d)
        
    def forward(self, x):
        #x = [sent len, batch size]
        embedded = self.dropout(self.embedding(x))
        #embedded = [sent len, batch size, emb dim]
        
        output, (hidden, cell) = self.rnn(embedded)
        
        #output = [sent len, batch size, hid dim * num directions]
        #hidden = [num layers * num directions, batch size, hid dim]
        #cell = [num layers * num directions, batch size, hid dim]
        
        #concat the final forward (hidden[-2,:,:]) 
        # and backward (hidden[-1,:,:]) hidden layers
        #and apply dropout
        
        hidden = self.dropout(torch.cat((hidden[-2,:,:], hidden[-1,:,:]), dim=1))
                
        #hidden = [batch size, hid dim * num directions]
            
        return self.fc(hidden.squeeze(0))
\end{lstlisting}

\begin{lstlisting}[language=Python, caption=Algorithm for Recurrent Neural Network AWD-LSTM applied to Text Data, numberstyle=\tiny]
from torch import *
class RNN_Encoder(nn.Module):

    """A custom RNN encoder network 
    """
     initrange=0.1
    def __init__(self, ntoken, emb_sz, nhid, nlayers, pad_token, bidir=False,
                 dropouth=0.3, dropouti=0.65, dropoute=0.1, wdrop=0.5):
        """ Default constructor for the RNN_Encoder class
            Returns:
                None
          """

        super().__init__()
        self.ndir = 2 if bidir else 1
        self.bs = 1
        self.encoder = nn.Embedding(ntoken, emb_sz, padding_idx=pad_token)
        self.encoder_with_dropout = EmbeddingDropout(self.encoder)
        self.rnns = [nn.LSTM(emb_sz if l == 0 else nhid,self.ndir,1,) for l in range(nlayers)]
        if wdrop: self.rnns = [WeightDrop(rnn, wdrop) for rnn in self.rnns]
        self.rnns = torch.nn.ModuleList(self.rnns)
        self.encoder.weight.data.uniform_(-self.initrange, self.initrange)

        self.emb_sz,self.nhid,self.nlayers,self.dropoute = emb_sz,nhid,nlayers,dropoute
        self.dropouti = LockedDropout(dropouti)
        self.dropouths = nn.ModuleList([LockedDropout(dropouth) for l in range(nlayers)])

    def forward(self, input):
        """ Invoked during the forward propagation of the RNN_Encoder module.
        Args:
            input (Tensor): input of shape (sentence length x batch_size)

        Returns:
            raw_outputs (tuple(list (Tensor), list(Tensor)): (no dropout, dropout)
        """
        sl,bs = input.size()
        if bs!=self.bs:
            self.bs=bs
            self.reset()

        emb = self.encoder_with_dropout(input, dropout=self.dropoute if self.training else 0)
        emb = self.dropouti(emb)

        raw_output = emb
        new_hidden,raw_outputs,outputs = [],[],[]
        for l, (rnn,drop) in enumerate(zip(self.rnns, self.dropouths)):
            current_input = raw_output
            with warnings.catch_warnings():
                warnings.simplefilter("ignore")
                raw_output, new_h = rnn(raw_output, self.hidden[l])
            new_hidden.append(new_h)
            raw_outputs.append(raw_output)
            if l != self.nlayers - 1: raw_output = drop(raw_output)
            outputs.append(raw_output)

        self.hidden = repackage_var(new_hidden)
        return raw_outputs, outputs

\end{lstlisting}
\section{Student: CNN}
Convolution Neural Networks have gained popularity as they perform at par with state of the art LSTM's networks and have the added advantage of of being faster at test time.
The CNN network for text classification differs from the convolutional networks used for computer vision \cite{szegedy2017inception,simonyan2014very}.
The architecture for classification uses only two or
three convolutional layers with large filters followed by more fully connected layers and only use convolution or pooling over one dimension, either time or frequency \cite{kim2014convolutional}. For the purpose our experiment we operate over time.
The embedding matrix is taken as input, keeping the embedding dimensions and the initialization same as that used by the teacher network.
Inspired by this we construct our CNN model (referred to as a-CNN) which will act as student learner.
We also use a deep CNN network \cite{conneau2016very} (referred as b-CNN), which works at the character level and performs small convolution and pooling operations as state of the art baseline of CNN architecture. 
The architecture of a-CNN is shown  in figure .
The Pseudo code of the class is provided below.
\begin{lstlisting}[language=Python, caption=Algorithm for Convolution Neural Network applied to Textual Data ]
import torch.nn as nn
import torch.nn.functional as F

class CNN(nn.Module):
   def __init__(self, v, emb_dim, n, f_sizes, o_dim, d):
       super().__init__()
       self.embbeding_dim = emd_dim
       self.embedding = nn.Embedding(v, self.embedding_dim)
       self.convs = nn.ModuleList([nn.Conv2d(in_channels=1,) out_channels=n,
       kernel_size=(fs,embedding_dim)) for fs in filter_sizes])
       self.fc = nn.Linear(len(filter_sizes)*n_filters, output_dim)
       self.dropout = nn.Dropout(dropout)
      
   def forward(self, x):
       #x = [sent len, batch size]
       x = x.permute(1, 0)
       #x = [batch size, sent len]
      
       embedded = self.embedding(x)
       #embedded = [batch size, sent len, emb dim]
      
       embedded = embedded.unsqueeze(1)
       #embedded = [batch size, 1, sent len, emb dim]
      
       conved = [F.relu(conv(embedded)).squeeze(3) for conv in self.convs]
       #conv_n = [batch size, n_filters, sent len - filter_sizes[n]]
      
       pooled = [F.max_pool1d(conv, conv.shape[2]).squeeze(2) for conv in conved]
       #pooled_n = [batch size, n_filters]
      
       cat = self.dropout(torch.cat(pooled, dim=1))
       #cat = [batch size, n_filters * len(filter_sizes)]
          
       return self.fc(cat)

\end{lstlisting}
\begin{figure}[h]

\includegraphics[scale=0.6]{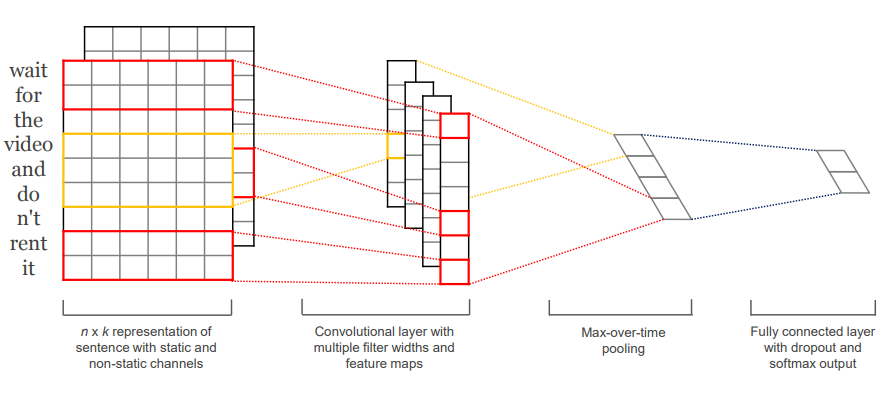}
\caption{Student Network} 
\end{figure}
\section{Blending}
Both RNNs and CNNs are powerful models, but the mechanisms that guide their learning are
quite different. That creates an opportunity to combine their predictions, implicitly averaging their
inductive biases. A classic way to perform this is ensembling, that is, to mix posterior predictions of
the two models in the following manner :
\begin{equation}
    p_{ensemble}(y|s) = \gamma p_{LSTM}(y|s) + (1 - \gamma) p_{CNN}(y|s)
\end{equation}
where  \(\gamma \in [0,1]\) and $s_{i} = (x_{1}, x_{2}, ...,x_{j}, ..., x_{m})$.
Also, $p_{LSTM}(y|s)$ and $p_{CNN}(y|s)$ are the probability of $y$ output class given a feature vector $x_{j}$ or sentence in this case.
This combination is in accordance with conditions proposed by \cite{dietterich2000ensemble} i.e.  \textit{“a necessary
and sufficient condition for an ensemble of classifiers to be more accurate than any of its individual
members is if the classifiers are accurate and diverse"} to form an ensemble of classifiers,but is inefficient at test time as it requires the posterior prediction of both the models at test time to compute the final predictions.
The method we propose is inspired by the process of knowledge distillation \cite{hinton2015distilling} which aims to distill knowledge from complex teacher models to much simpler student models by matching output logits, as well as model blending \cite{geras2015blending} which uses the posterior predictions of LSTM to train vision style CNN architectures for automatic speech recognition task.
We use a different class of CNN which utilizes large filter sizes and are shallow in comparison to the vision style CNN.
We train our CNN architecture for text classification task namely sentiment analysis, question classification and topic classification.
To the best of our knowledge, this is first time CNN architectures are trained in the presence of LSTM priors for these classification tasks.
To incorporate the LSTM information during training of the CNN we modify the loss function to combine loss from  hard labels from the training data with a loss function which penalises deviation from predictions of the LSTM teacher.
The training objective is to minimize the weighted sum of the cross-entropy loss :

\begin{equation}
\mathcal{L}= \lambda\cdot \mathcal{L}_{soft}+{1-\lambda}
\cdot \mathcal{L}_{hard} 
\end{equation}
\begin{equation}
\mathcal{L}_{soft} = H (p_{LSTM}(c|s_{i}),q_{CNN}(c|s_{i}))
\end{equation}
\begin{equation}
\mathcal{L}_{hard} = H(q_{CNN}(y_{i}|s_{i}), y_{i}^{true})
\end{equation}

where $p_{LSTM}(c|s_{i})$ is the probability of class c for the the given training example $s_{i}$ predicted by the teacher LSTM network, $q_{CNN}(c|s_{i})$ is the probability of class c for the the given training example $s_{i}$ predicted by the student CNN and $y_{i}^{true}$ is the true class for the training example $s_{i}$.
Here $\lambda \in [0,1]$, helps us control the relative contribution of the two losses.
This modified loss function helps us to incorporate information learnt by LSTM network while training the student CNN network, this method give similar performance to the ensemble of the two models, but it is computationally 17x less expensive at test time than the ensemble.
Thus blending results in more accurate and efficient classifiers at test time.

%% file: Chapters/Chapter4.tex

\chapter{Experiments and Results} 

\label{Chapter 4} 


\lhead{Chapter 4. \emph{Experiments}} 

\section{Dataset}
We evaluate our models on four public dataset namely IMDB dataset \cite{maas-EtAl:2011:ACL-HLT2011} for sentiment classification, TREC-6 \cite{voorhees1999trec} for question classification, AG's corpus of news articles for topic modeling and DBpedia by \cite{zhang2015character}.
The description of the datasets are tabulated below:
\renewcommand{\arraystretch}{1.2}
\begin{table}
        \caption{Description of datasets}
        \centering
\begin{tabular}{ c|c|c|c|c}
 \hline
Dataset & IMDB & TREC-6 & AG & DBPedia  \\
 \hline
Training & 25k & 5452 & 120k & 560k\\
Testing & 25k & 500 & 7600 & 70k\\
Classes & 2 & 6 & 4 & 14\\
Avg-words & 234 & - & 45 & 55\\
 \hline
\end{tabular}
\end{table}
\begin{table}
        \caption{Comparision across datasets. one column for each dataset is accuracy }
        \centering
\begin{tabular}{ c|c|c|c|c|c}
 \hline
\multicolumn{2}{c|}{Model} &  {IMDB} & {TREC-6} & {AG} & {DBPedia}  \\
\hline
\multirow{1}{4em}{Linear}  & Fasttext  \cite{joulin2016bag} &   85.22  &  88.2  &   92.6  &   98.5  \\ 
 \hline
 \multirow{3}{4em}{RNN}  & LSTM &   82.77  & 90.2 &   82.77  & - \\ 
 & Bi-LSTM \cite{zhou2016text} &   84.11  & 93.0 &   93.0  & 99.12 \\
 & AWD-LSTM \cite{merity2017regularizing} &   95.4 & 96.4 &   95.0  & 99.20\\
 \hline
 \multirow{3}{4em}{CNN}  & a-CNN \cite{kim2014convolutional} & 88.22  & 92.0 &   91.6   & 98.6 \\
 & b-CNN \cite{conneau2016very}&   -  & 93.0 &   91.3  & 98.4  \\
 & c-CNN \cite{zhang2015character}&   -  & 93.2 &   87.2 (90.49) & 98.3  \\
 \hline 
  \multirow{2}{4em}{Ensemble}  & CNN + LSTM &   88.63  & 93.15 & 92.0 & 99.0 \\
 & CNN + AWD-LSTM &   95.4  & 96.7 &   95.0 & 99.23 \\
 \hline
  \multirow{2}{4em}{Blended*}  & CNN + LSTM &   87.23  & 91.08 &   92.3 & 99.14 \\
  & CNN + AWD-LSTM &   93.2 &  92.34  &  92.5 & 99.1 \\
 \hline
\end{tabular}
\end{table}
\begin{table}
        \caption{Comparison on the execution time of the methods is execution time compared to the our proposed CNN \protect\cite{kim2014convolutional} on IMDB dataset.}
        \centering
\begin{tabular}{ c|c|c}
 \hline
\multicolumn{2}{c|}{Model} &  {Execution Time}  \\
\hline
\multirow{1}{4em}{Linear}  & Fasttext &   1.1x   \\ 
 \hline
 \multirow{3}{4em}{RNN}  & LSTM & 5.55x  \\ 
 & Bi-LSTM &   5.64x \\
 & AWD-LSTM &   15.5x \\
 \hline
 \multirow{3}{4em}{CNN}  & a-CNN & 1.0x \\
 & b-CNN &   4.3x  \\
 \hline 
  \multirow{2}{4em}{Ensemble}  & a-CNN + LSTM &   6.2x   \\
 & a-CNN + AWD-LSTM &   16.7x   \\
 \hline
  \multirow{2}{4em}{Blended*}  & a-CNN + LSTM &   1.0x \\
  & a-CNN + AWD-LSTM & 1.0x  \\
 \hline
\end{tabular}
\end{table}

\subsection{Implementation Details}

\textbf{Input.} We adopted the word vectors from \cite{pennington2014glove}.The word embedding used were of size 300 for our LSTM network. 
\newline
\textbf{Training Setting.}
For training the state of the art AWD-LSTM \cite{merity2017regularizing} Teacher we follow similar training regime as \cite{howard2018universal} with an embedding size of 400, 4 layers,1150 hidden activations per layer, and a BPTT batch size of 70. 
We apply different dropout to different types as suggested by and  \cite{merity2017regularizing}
The student CNN network \cite{kim2014convolutional,zhang2015character} has filter sizes of 3,4 and 5.
The model were implemented in pytorch and were adapted from prior works \cite{howard2018universal,kim2014convolutional}.
The CNN trained using the proposed method outperforms the CNN architecture and the teacher network and is 15x more computationally efficient from its teacher.
The experiments were first performed on IMDB dataset for sentiment analysis to tune $\gamma$ and $\lambda$ hyper-parameter to their optimal values of 0.4 and 0.5 we then validate these parameter using the TREC-6 data set for question classification.
We use Adam with default setting $\beta_{1} = 0.9 $ and $\beta_{2} =0.99$ with a minibatch of 64.The practises otherwise are same as used by \cite{howard2018universal,merity2017regularizing}

\subsection{Baseline and Results}
We compare our method and utilize several of the popular models in our experiments : linear model fasttext \cite{joulin2016bag}, we use CNN architecture inspired by \cite{kim2014convolutional} and train it using numerous teacher \cite{merity2017regularizing,zhou2016text} which are state of the art for text classification.
We compare the model with character level CNN \cite{zhang2015character} referred to as c-CNN and very Deep CNN network \cite{conneau2016very}.
We also benchmark it against \cite{Wang2018Densely} which proposes dense connection between convolutional layer and multi-scale feature attention for text classification.
They also experiment with depth of the network an found that shallow networks perform comparable to the deep models with only a slight drop in performance.
Shallowness of the model is advantageous for deployment to low - resource devices. 
Our proposed training regime improves the performance of shallow CNN and performs comparable to other popular networks of the type.

\section{Main Results}
The CNN trained through the proposed method is comparable to its Teacher LSTM network, due to the shallowness of the models it is more interpretable and is faster at the test time and hence the model can be deployed to mobile devices.
The provides evidence towards using a trained complex network transferring the 'dark knowledge' \cite{hinton2015distilling} to a much simpler model which is computationally efficient at inference time.

%% file: Chapters/Chapter5.tex

\chapter{Conclusion} 

\label{Chapter 5} 


\lhead{Chapter 5. \emph{Conclusion}} 
We see that blending the state of the art RNN into the CNN
model, results in minimal drop in accuracy . But the inference time is reduced
greatly, by about 18 times compared to the ensemble. This shows that exploring blending of state of the art
RNN models into CNN models is promising, and much more can be achieved if we
sacrifice a bit on the inference time (by making the CNN model more deep, for
example).\textbf{This also increase deployability of these models to mobile devices i.e. Smartphones, Cameras etc.}

%% file: Chapters/Chapter6.tex

\chapter {Future Work} 

\label{Chapter 6} 


\lhead{Chapter 6. \emph{Future Work}} 
The current CNN architecture we used had only a single convolutional layer. We can try blending into CNN architectures with more than one
layer. This would increase the inference time a bit, but the accuracy would also
improve. Basically, we try experimenting with the CNN architecture until we reach
a sweet spot with respect to both inference time and accuracy. Moreover, our current work deals with only sentence classification. We can try to apply this technique
to other natural language tasks which require sequence modelling such as machine
translation, in which the state of the art models are RNN based.

%% file: Appendices/AppendixA.tex
\chapter{Appendix A} 

\label{AppendixX} 

\lhead{Appendix X. \emph{Appendix Title Here}} 

This contains basic implementation of the above mentioned algorithm in python it has the following dependencies which are mentioned below:

\begin{itemize}
    \item \textbf{$Python >= 3.0$}
    \item \textbf{$fastai == 0.7.0$}
    \item \textbf{$torchtext == 0.2.3$}
\end{itemize}

The experiments were performed on \textbf{Google COLAB}, which is free jupyter notebook service which has GPU capabilities.Experiments were performed on \textbf{NVIDIA K80 GPU}, and testing of the models were done on \textbf{MI A2 mobile device} by converting the \textbf{Pytorch model to Tensorflow model using ONYX.}

%% file: main.bbl
\begin{thebibliography}{10}
\providecommand{\url}[1]{\texttt{#1}}
\providecommand{\urlprefix}{URL }
\providecommand{\doi}[1]{https://doi.org/#1}

\bibitem{ashok2017n2n}
Ashok, A., Rhinehart, N., Beainy, F., Kitani, K.M.: N2n learning: network to
  network compression via policy gradient reinforcement learning. arXiv
  preprint arXiv:1709.06030  (2017)

\bibitem{bai2018empirical}
Bai, S., Kolter, J.Z., Koltun, V.: An empirical evaluation of generic
  convolutional and recurrent networks for sequence modeling. arXiv preprint
  arXiv:1803.01271  (2018)

\bibitem{conneau2016very}
Conneau, A., Schwenk, H., Barrault, L., Lecun, Y.: Very deep convolutional
  networks for text classification. arXiv preprint arXiv:1606.01781  (2016)

\bibitem{dauphin2016language}
Dauphin, Y.N., Fan, A., Auli, M., Grangier, D.: Language modeling with gated
  convolutional networks. arXiv preprint arXiv:1612.08083  (2016)

\bibitem{dietterich2000ensemble}
Dietterich, T.G.: Ensemble methods in machine learning. In: International
  workshop on multiple classifier systems. pp. 1--15. Springer (2000)

\bibitem{gehring2016convolutional}
Gehring, J., Auli, M., Grangier, D., Dauphin, Y.N.: A convolutional encoder
  model for neural machine translation. arXiv preprint arXiv:1611.02344  (2016)

\bibitem{geras2015blending}
Geras, K.J., Mohamed, A.r., Caruana, R., Urban, G., Wang, S., Aslan, O.,
  Philipose, M., Richardson, M., Sutton, C.: Blending lstms into cnns. arXiv
  preprint arXiv:1511.06433  (2015)

\bibitem{graves2013generating}
Graves, A.: Generating sequences with recurrent neural networks. arXiv preprint
  arXiv:1308.0850  (2013)

\bibitem{graves2009offline}
Graves, A., Schmidhuber, J.: Offline handwriting recognition with
  multidimensional recurrent neural networks. In: Advances in neural
  information processing systems. pp. 545--552 (2009)

\bibitem{han2015deep}
Han, S., Mao, H., Dally, W.J.: Deep compression: Compressing deep neural
  networks with pruning, trained quantization and huffman coding. arXiv
  preprint arXiv:1510.00149  (2015)

\bibitem{hinton2015distilling}
Hinton, G., Vinyals, O., Dean, J.: Distilling the knowledge in a neural
  network. arXiv preprint arXiv:1503.02531  (2015)

\bibitem{howard2018universal}
Howard, J., Ruder, S.: Universal language model fine-tuning for text
  classification. In: Proceedings of the 56th Annual Meeting of the Association
  for Computational Linguistics (Volume 1: Long Papers). vol.~1, pp. 328--339
  (2018)

\bibitem{joulin2016bag}
Joulin, A., Grave, E., Bojanowski, P., Mikolov, T.: Bag of tricks for efficient
  text classification. arXiv preprint arXiv:1607.01759  (2016)

\bibitem{kalchbrenner2016neural}
Kalchbrenner, N., Espeholt, L., Simonyan, K., Oord, A.v.d., Graves, A.,
  Kavukcuoglu, K.: Neural machine translation in linear time. arXiv preprint
  arXiv:1610.10099  (2016)

\bibitem{kim2014convolutional}
Kim, Y.: Convolutional neural networks for sentence classification. arXiv
  preprint arXiv:1408.5882  (2014)

\bibitem{maas-EtAl:2011:ACL-HLT2011}
Maas, A.L., Daly, R.E., Pham, P.T., Huang, D., Ng, A.Y., Potts, C.: Learning
  word vectors for sentiment analysis. In: Proceedings of the 49th Annual
  Meeting of the Association for Computational Linguistics: Human Language
  Technologies. pp. 142--150. Association for Computational Linguistics,
  Portland, Oregon, USA (June 2011),
  \url{http://www.aclweb.org/anthology/P11-1015}

\bibitem{merity2017regularizing}
Merity, S., Keskar, N.S., Socher, R.: Regularizing and optimizing lstm language
  models. arXiv preprint arXiv:1708.02182  (2017)

\bibitem{oord2016wavenet}
Oord, A.v.d., Dieleman, S., Zen, H., Simonyan, K., Vinyals, O., Graves, A.,
  Kalchbrenner, N., Senior, A., Kavukcuoglu, K.: Wavenet: A generative model
  for raw audio. arXiv preprint arXiv:1609.03499  (2016)

\bibitem{pennington2014glove}
Pennington, J., Socher, R., Manning, C.D.: Glove: Global vectors for word
  representation. In: Empirical Methods in Natural Language Processing (EMNLP).
  pp. 1532--1543 (2014), \url{http://www.aclweb.org/anthology/D14-1162}

\bibitem{polino2018model}
Polino, A., Pascanu, R., Alistarh, D.: Model compression via distillation and
  quantization. arXiv preprint arXiv:1802.05668  (2018)

\bibitem{simonyan2014very}
Simonyan, K., Zisserman, A.: Very deep convolutional networks for large-scale
  image recognition. arXiv preprint arXiv:1409.1556  (2014)

\bibitem{szegedy2017inception}
Szegedy, C., Ioffe, S., Vanhoucke, V., Alemi, A.A.: Inception-v4,
  inception-resnet and the impact of residual connections on learning. In:
  AAAI. vol.~4, p.~12 (2017)

\bibitem{vinyals2015grammar}
Vinyals, O., Kaiser, {\L}., Koo, T., Petrov, S., Sutskever, I., Hinton, G.:
  Grammar as a foreign language. In: Advances in Neural Information Processing
  Systems. pp. 2773--2781 (2015)

\bibitem{voorhees1999trec}
Voorhees, E.M., Tice, D.M.: The trec-8 question answering track evaluation. In:
  TREC. vol.~1999, p.~82 (1999)

\bibitem{Wang2018Densely}
Wang, S., Huang, M., Deng, Z.: Densely connected cnn with multi-scale feature
  attention for text classification  (2018)

\bibitem{wang2012baselines}
Wang, S., Manning, C.D.: Baselines and bigrams: Simple, good sentiment and
  topic classification. In: Proceedings of the 50th Annual Meeting of the
  Association for Computational Linguistics: Short Papers-Volume 2. pp. 90--94.
  Association for Computational Linguistics (2012)

\bibitem{zhang2015character}
Zhang, X., Zhao, J., LeCun, Y.: Character-level convolutional networks for text
  classification. In: Advances in neural information processing systems. pp.
  649--657 (2015)

\bibitem{zhou2016text}
Zhou, P., Qi, Z., Zheng, S., Xu, J., Bao, H., Xu, B.: Text classification
  improved by integrating bidirectional lstm with two-dimensional max pooling.
  arXiv preprint arXiv:1611.06639  (2016)

\bibitem{zhu2016trained}
Zhu, C., Han, S., Mao, H., Dally, W.J.: Trained ternary quantization. arXiv
  preprint arXiv:1612.01064  (2016)

\end{thebibliography}
